\documentclass[10pt,twocolumn,letterpaper]{article}

\usepackage{iccv}
\usepackage{times}
\usepackage{epsfig}
\usepackage{graphicx}
\usepackage{amsmath}
\usepackage{amssymb}
\usepackage{algorithm,algorithmic}
\usepackage{multirow}
\usepackage[dvipsnames]{xcolor}
\usepackage{xr}
\usepackage{authblk}
\usepackage[pagebackref=true,breaklinks=true,letterpaper=true,colorlinks,bookmarks=false]{hyperref}

\iccvfinalcopy 

\begin{document}

\title{RFLA: A Stealthy Reflected Light Adversarial Attack in the Physical World}

\author[1]{Donghua Wang}
\author[2]{Wen Yao\thanks{Corresponding Author.}}
\author[2]{Tingsong Jiang$^{\small*}$}
\author[3]{Chao Li}
\author[2]{Xiaoqian Chen}
\affil[1]{College of Computer Science and Technology, Zhejiang University}
\affil[2]{Defense Innovation Institute, Chinese Academy of Military Science}
\affil[3]{School of Artificial Intelligence, Xidian University}
\affil[ ]{wangdonghua@zju.edu.cn, \{wendy0782,lichaoedu\}@126.com, tingsong@pku.edu.cn, chenxiaoqian@nudt.edu.cn}

\maketitle

\begin{abstract}
Physical adversarial attacks against deep neural networks (DNNs) have recently gained increasing attention. The current mainstream physical attacks use printed adversarial patches or camouflage to alter the appearance of the target object. However, these approaches generate conspicuous adversarial patterns that show poor stealthiness. Another physical deployable attack is the optical attack, featuring stealthiness while exhibiting weakly in the daytime with sunlight. In this paper, we propose a novel Reflected Light Attack (RFLA), featuring effective and stealthy in both the digital and physical world, which is implemented by placing the color transparent plastic sheet and a paper cut of a specific shape in front of the mirror to create different colored geometries on the target object. To achieve these goals, we devise a general framework based on the circle to model the reflected light on the target object. Specifically, we optimize a circle (composed of a coordinate and radius) to carry various geometrical shapes determined by the optimized angle. The fill color of the geometry shape and its corresponding transparency are also optimized. We extensively evaluate the effectiveness of RFLA on different datasets and models. Experiment results suggest that the proposed method achieves over 99\% success rate on different datasets and models in the digital world. Additionally, we verify the effectiveness of the proposed method in different physical environments by using sunlight or a flashlight.
\end{abstract}

\section{Introduction}

\begin{figure}[ht]
	\centering
	\begin{minipage}{1.\linewidth}
		\centering
		\includegraphics[width =1.\linewidth]{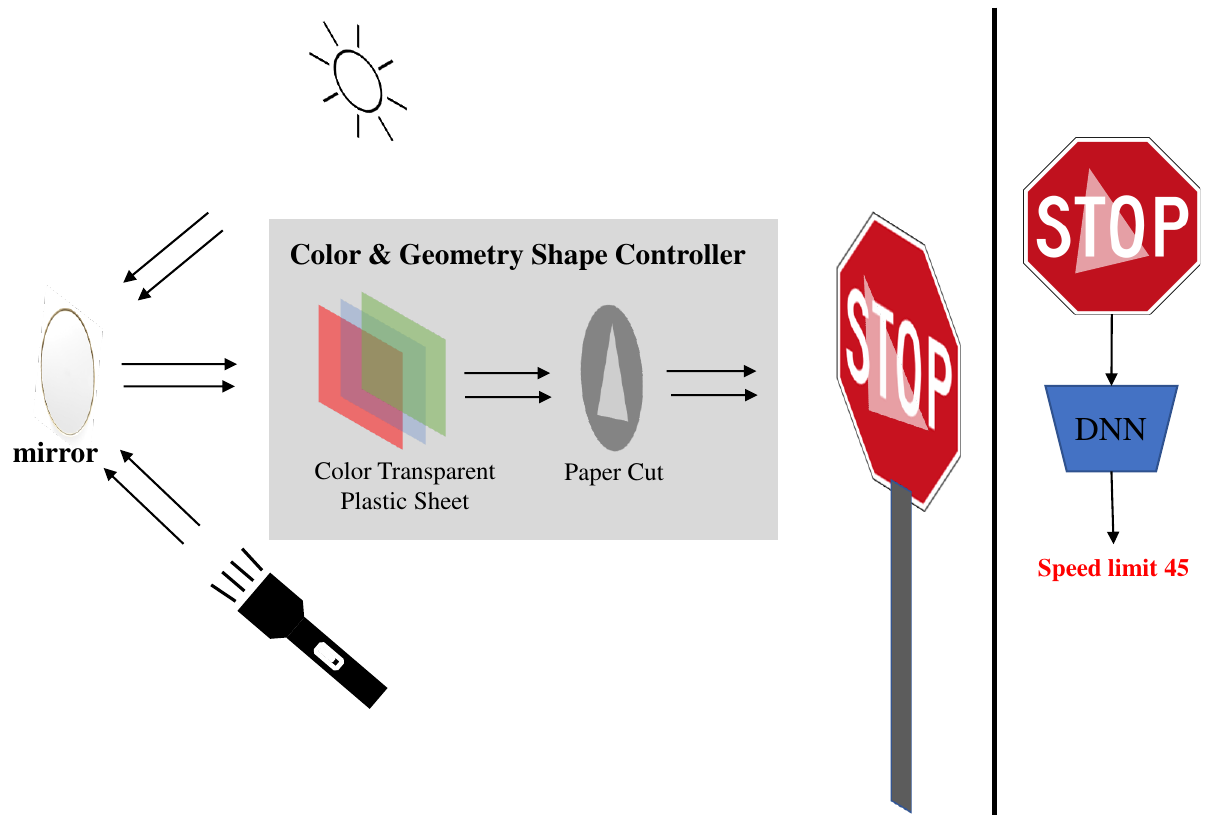}
	\end{minipage}
\caption{The reflected light is modulated by the color transparency plastic sheet and paper cut of the specific shape for better attack performance. The Reflected light source can be sunlight or a flashlight (when the sunlight is unreachable).}
\label{fig:tile_figure}
\end{figure}

Deep neural networks (DNNs) have increasingly been applied to daily life as their dramatic capabilities, such as automatic driving, facial payment, and computer-aided diagnosis. However, DNN-based systems have exposed security risks caused by adversarial examples \cite{szegedy2014intriguing}. Adversarial examples are crafted by carefully designed noise that is invisible to humans but can deceive the DNNs. Furthermore, recent researches \cite{sharif2016accessorize,ifgsm2018adversarial} reported that physically deployed DNN-based systems are also exposed to such security risks. Therefore, exploring various potential risks in security-sensitive systems to avoid possible loss is urgent.

Existing adversarial attack methods can be categorized into digital attacks and physical attacks. The former focus on pursuing higher attack performance on limitation conditions, such as breaking the model equipped with adversarial defense \cite{bim2016adversarial,carlini2017towards,tifgsm2019evading,sifgsm2019nesterov}, preventing the attacker from accessing the target model's information (e.g., architecture or dataset), i.e., black-box attack \cite{alzantot2019genattack,li2022approximate,li2023adaptive}. Although some researchers suggested that adversarial examples generated by the digital attack can be applied to physical attacks \cite{ifgsm2018adversarial}, the attack performance is not satisfying. The possible reason is that the adversarial perturbation is too small to resist the environmental noise in the physical world. In contrast, physical attacks are designed to be physically deployable, where one crucial change is eliminating the perturbation's magnitude constraint.

A line of physical adversarial attack methods \cite{eykholt2018robust,duan2022learning,wang2022fca,wang2021dual} has been proposed, which can be grouped into contact attacks and contactless attacks. The former requires the attacker to approach the target object and then modify the appearance of the target object by pasting the adversarial patch or camouflage. However, the adversarial pattern generated by these methods is conspicuous, which easily alerts humans and results in attack failure. By contrast, contactless physical attacks do not require the attacker to approach the target object while modifying the appearance of the target object by projecting or emitting light or laser on the target object to perform attacks, making it stealthy and dangerous. Optical attacks are representative contactless attacks. Although several optical attacks have been proposed \cite{man2019poster,sayles2021invisible,duan2021adversarial,huang2022spaa}, they merely work in dark environments as the strong light (e.g., sunlight) environment would disturb the emitted light, limiting their usability.

In this paper, we get inspiration from the fact that the driver is easily affected by the strong reflected light, resulting in a potential car accident, and such potential risks to the automatic driving system remain unexplored. We explore the vulnerability of the DNNs toward the reflected light attack by elaborately designing the position, geometry, and color of the reflected light. Specifically, we propose a Reflected Light Attack (RFLA), which can solve the issue of the poor attack performance of existing optical attacks in strong-light environments, as our light source is sunlight. To perform physical attacks, we use a mirror to reflect the sunlight toward the target object to modify its appearance. However, the monotonous sunlight (usually white) may not obtain the desired performance. Therefore, we first use different colored transparent plastic sheets to modulate the color of the reflected light, then apply a paper cut of a specific shape to control the shape of reflected light on the target object (see Figure \ref{fig:tile_figure}). Finally, we can create different colors and shapes of the reflected light on the target object's specific region to achieve desired attack performance. 

To achieve the above goals, we present a general framework based on the circle to model the above problem. Specifically, we first initialize a circle with a random coordinate and radius. On this circle, we create a point on the circle using sine and cosine with a randomly selected angle. Then, we customize a shape by adding a new angle, which is used to create a new point in the circumference. The other points required to create a geometry can be obtained by applying the center symmetry of the circle. Moreover, the fill color and its transparency are also considered in the optimization. Finally, we adopt the particle swarm optimization (PSO) algorithm to find the optimal result.
Our contributions are listed as follows.
\begin{itemize}
\item We propose a novel reflect-light-based physical adversarial attack under the black-box scenario. It reflects the natural sunlight toward the target object using a mirror, making it controllable and stealthy.
\item We devise a general framework based on a circle to search for the best position, geometry, and color of the reflected light to achieve better attack performance. 
\item We comprehensively investigate the influence of the geometry, position, and color of the reflected light on attack performance in the digital world. We conduct the physical adversarial attack by using sunlight for daytime and a flashlight for sunlight unavailable, and the experiment results verify the effectiveness of the proposed method.  
\end{itemize}

\section{Related Works}

\subsection{Digital adversarial attacks}
Digital adversarial attacks have enjoyed decade development, which can be roughly divided into white-box attack methods and black-box attack methods. The former grants the adversary access to the target model, allowing them to develop attack algorithms with the model's gradient. The most represented gradient-based attack is the fast gradient sign method (i.e., FGSM \cite{fgsm2015explaining}), which updates adversarial examples along the ascending direction of the gradient under one iteration step. Since then, a line of variants has been proposed, including an iterative variant of FGSM (i.e., I-FGSM \cite{ifgsm2018adversarial}), random initialization has been adopted (i.e., PGD \cite{pgd2018towards}), momentum term is introduced to enhance the transferability (i.e., MI-FGSM \cite{mim2018BoostingAA}), and various data augmentation technique like diversity input (i.e., DI-FGSM\cite{difgsm2019improving}), translation-invariant (i.e., TI-FGSM \cite{tifgsm2019evading}) and scale-invariant (i.e., SI-FGSM \cite{sifgsm2019nesterov}). In contrast, black-box attacks prohibit the attacker from accessing any information about the target model but are open for queries, which makes black-box attacks more challenging. Nonetheless, many black-box attacks are proposed, such as exploiting the differential evolution algorithm \cite{li2022approximate}, genetic algorithm \cite{alzantot2019genattack}, particle swarm optimization, and so on \cite{andriushchenko2020square}. In addition, several works suggested that adversarial perturbation's position \cite{wei2022adversarial}, pattern \cite{yang2020patchattack}, and geometry \cite{chen2022shape} on the clean image significantly impact attack performance. However, current works only investigate one or two of these factors. In this work, we systematically investigate the influence of the adversarial perturbation's position, geometry, and pattern on attack performance under the black-box scenario.

\subsection{Physical adversarial attacks}
According to whether it requires the attacker to access the target object in the real attack scenario, physical adversarial attacks can be grouped into contact and contactless physical attacks. Contact attacks can be further categorized into patch-based attacks and camouflage-based attacks. Patch-based attacks mainly focus on optimizing an adversarial image patch, which is then printed out and stuck on the target object or held by the attacker to deceive the target DNNs. Patch-based attacks are usually applied in attacking the facial recognition model \cite{sharif2016accessorize,sharif2019general,wei2022adversarial,wei2022simultaneously}, pedestrian detection model \cite{thys2019fooling,tan2021legitimate,hu2021naturalistic,doan2022tnt}, and traffic sign recognition model \cite{eykholt2018robust,liu2019perceptual,zhong2022shadows}. Camouflage-based attacks \cite{wang2021dual,wang2022fca,duan2022learning} slightly differ from patch-based, as they concentrate on modifying the appearance of the target object via UV Texture. Thus, camouflage-based attacks show better attack performance in the multi-view scenario by painting the full coverage camouflage over the appearance of the target object. However, although contact physical attacks achieve good physical attack performance, the pattern of the adversarial patch/camouflage is conspicuous, which leads to poor stealthiness.
In contrast, contactless physical attacks are performed by projecting/emitting light \cite{gnanasambandam2021optical,huang2022spaa}, or a laser beam \cite{duan2021adversarial}, usually called optical attacks. However, existing optical attacks work in dark environments \cite{duan2021adversarial,huang2022spaa} while performing poorly in strong-light environments. The reason is that the light beam emitted by a light source is easily affected by environmental light, resulting in attacking failure. Recently, Zhang \textit{et al.} \cite{zhong2022shadows} proposed a shadow-based attack, but it can only create a triangle shape with one monotonous color (e.g., gray). In this work, we solve the situation of poor attacks in strong-light (i.e., sunlight) environments to perform attacks since we directly use sunlight to perform attacks. Moreover, we create reflected light with different geometrical shapes and colors using the color transparency plastic sheet and paper cut.


\section{Methodology}

\subsection{Problem statement}
Let $\mathcal{X}$ denote the data distribution, and the corresponding ground truth label is $\mathcal{Y}$. Given an image $x \in \mathcal{X}$ that has the resolution of $x \in \mathbb{R}^{C \times H \times W}$, a well trained neural network $f$ output $\hat{y} = f(x)$ and $\hat{y} = y$, where the $\hat{y}$ is prediction of the $f$ and $y$ is the ground truth label, $\hat{y}, y \in \mathbb{R}^{|\mathcal{Y}|}$. Adversarial attack aims to generate adversarial examples $x_{adv}$ to make the $f$ output the wrong prediction by adding small perturbation $\delta$ into the clean image $x$, i.e., $x_{adv} = x + \delta$. Mathematically, the $\delta$ is obtained by solving the following problem

\begin{equation}
\min~~\delta ~~~ s.t.~~ f(x+\delta) \neq f(x), ~ ||\delta||_p \leq \epsilon,
\label{eq:adv_noraml}
\end{equation}
where $||\cdot||_p$ is the $L_p$ norm, which bound the maximum allowable magnitude of $\delta$. 

The optimization objective of Equation \ref{eq:adv_noraml} is the general formal for constructing the full-pixel-wise perturbation, which is unsuitable for physical adversarial attacks as the background of the physical world is unchangeable. Therefore, we reformulate Equation \ref{eq:adv_noraml} to optimize the physical deployable perturbation by modifying the construction of $x_{adv}$. Specifically, we define an apply function $\mathcal{A}(x, p, l, M)$ to construct adversarial examples $x_{adv}$, which indicates that apply the perturbation $p$ at the location $l$ of the clean image $x$, where $M$ is the binary mask to indicate whether the position is allowed to modify (one denote allow while zero not). 

In this work, we aim to reflect the sunlight toward the target object to perform stealthy physical adversarial attacks, where the representation (e.g., geometry, fill color and position) of reflected light on the target object is the key to a successful attack. Therefore, the parameters of $\mathcal{A}(x, p, l, M)$ comprise geometry and fill color of $p$, and the location of $l$ are variables to be optimized.

\subsection{Reflected light Attack} \label{sec:reflect_attack}
Sunlight is the most common and indispensable natural phenomenon in daily life. People can reflect the sunlight toward the wall to construct various shapes by using different shapes of mirrors. However, the danger of such reflected light against DNN-based systems has been ignored, which may pose a potential risk as it featured extremely stealthy and controllable. In this work, we aim to modulate the reflected light to perform adversarial attacks in the digital and physical world.

\begin{figure}[t]
	\centering
	\begin{minipage}{.3\linewidth}
		\centering
		\includegraphics[width =1.\linewidth]{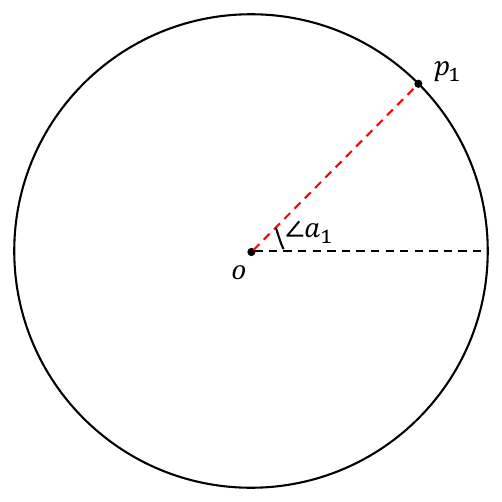}
		\centerline{\footnotesize (a) Base circle}
	\end{minipage}
	\begin{minipage}{.3\linewidth}
		\centering
		\includegraphics[width =1.\linewidth]{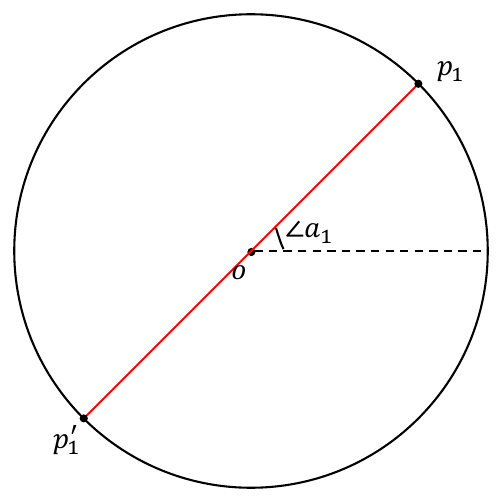}
		\centerline{\footnotesize (b) Line}
	\end{minipage}
	\begin{minipage}{.3\linewidth}
		\centering
		\includegraphics[width =1.\linewidth]{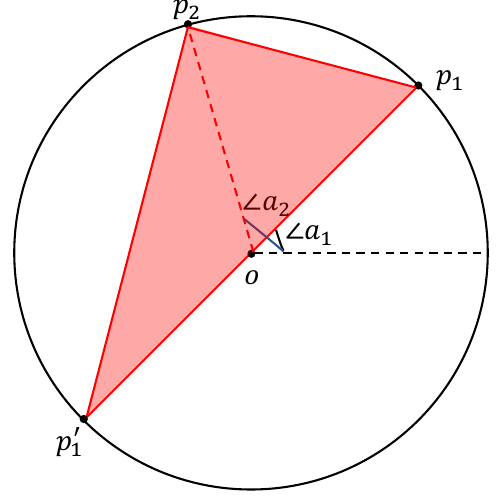}
		\centerline{\footnotesize (c) Triangle}
	\end{minipage}
	
	\begin{minipage}{.3\linewidth}
		\centering
		\includegraphics[width =1.\linewidth]{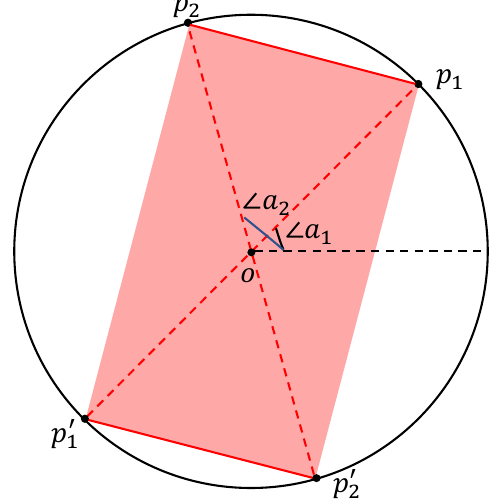}
		\centerline{\footnotesize (d) Rectangle}
	\end{minipage}
	\begin{minipage}{.3\linewidth}
		\centering
		\includegraphics[width =1.\linewidth]{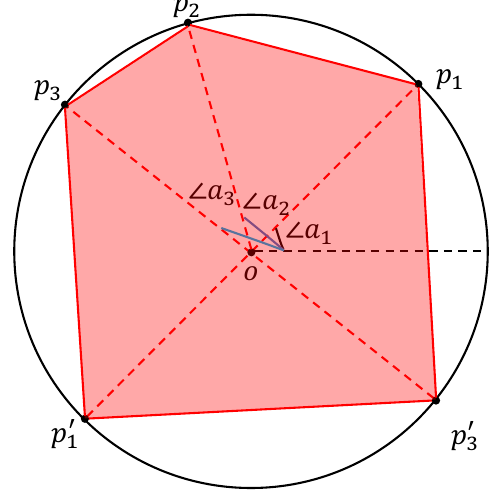}
		\centerline{\footnotesize (e) Pentagon}
	\end{minipage}
	\begin{minipage}{.3\linewidth}
		\centering
		\includegraphics[width =1.\linewidth]{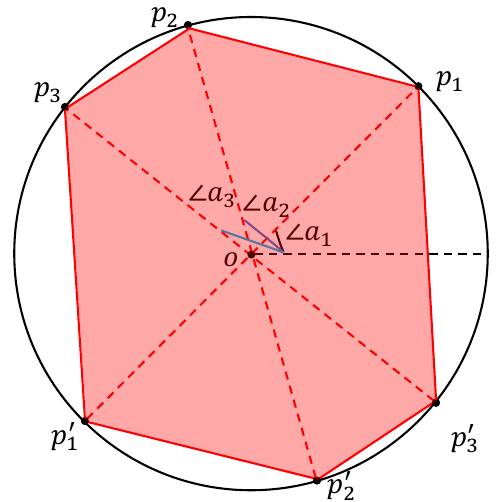}
		\centerline{\footnotesize (f) Hexagon}
	\end{minipage}
\caption{Visualization explanation of the circle modeling. Different geometries are constructed by adjusting the angle. For example, the triangle is created by adding a new angle $\angle {a_2}$ on the base circle (a), where three points are composed of two points ($p_1$ and $p_2$) determined by two angles ($\angle {a_1}$ and $\angle {a_2}$) and one symmetry point ($p'_1$).}
\label{fig:reflect_attack}
\end{figure}

Previous work \cite{zhong2022shadows} modeled the triangle shadow by optimizing three points, which requires complex constraints on points to construct a rational geometry shape if they extend to other geometric shapes. To address this issue, we exploit the characteristic of the circle and propose a novel general framework based on the circle, which can generate various shapes by adjusting the number of angles (see Figure \ref{fig:reflect_attack}). The details process is described as follows. 

\begin{itemize}
\item Select a radius $r$ from the region of $[0, min(H, W)/2]$.
\item Spawn a center $o(x, y)$ of the circle from the region of $[r, H-r]$ and $[r, W-r]$.
\item Randomly select an angle $a_1$ and spawn a point $p_1(x_1,y_1)$ on circle by the follow equation 
\begin{equation}
\left\{
\begin{array}{lr}
x_1 = x + r\times \sin (\frac{a_1 \times \pi}{180}), & \\
y_1 = y + r\times \cos (\frac{a_1 \times \pi}{180}), &
\end{array}
\right.
\end{equation}
\item Calculate the symmetry point $p'_1(x'_1,y'_1)$ of the point $p_1$ against the center of the circle by $x'_1 = 2\times x - x_1$ and $y'_1 = 2\times y - y_1$.
\item Randomly select a color tuple $(red, green, blue)$ from the region of [0, 255], and the transparency $\alpha$ from [0, 1].
\end{itemize}

The above process can plot a line on the clean image. To construct varying geometries like a triangle, rectangle, pentagon, or hexagon, one can repeat the third and fourth steps to create a new point by adding a new angle. Algorithm \ref{alg:gen_pop} in describes the detailed particle initialization process.

\subsection{Optimization}
As aforementioned, we have eight base variables that need to be optimized, expressed as eight-tuples $(x, y, r, \alpha, red, green, blue, a_1)$, which can be used to plot a line on the clean image. To generate various geometries, more additional variables are required to generate various geometries, which depend on the shape to be generated. For example, there is one extra variable for the triangle and rectangle; two extra for the pentagon and hexagon. Note that the proposed method is easily extended to more complex geometry. Recall that our goal is to deceive the DNNs by plotting geometry on the clean image. Thus, we adopt the particle swarm optimization algorithm (PSO) to seek the best geometry, fill color, and position.

In PSO, we represent the optimization variables tuple as a particle (i.e., the solution vector $q$). The update direction of the particle is determined by a velocity vector $v$. Every particle stands for a potential solution, which requires to be optimized. We treat the personal historical best solution of a particle as $q_{pbest}$, and the global best solution of a particle as $q_{gbest}$. Moreover, for every solution, we fix the circle's position and merely optimize the geometry, fill color, and transparency. Therefore, to represent the best solution of a circle, we devised an additional metric $q_{sgbest}$, which is the sum of the fitness score of all geometrical shapes in a specific circle. Finally, the update criterion is defined as follows:

\begin{equation}
\begin{split}
v_i(t) =  Wv_i(t-1) + C_1\kappa_1(q_{pbest} - q_i(t)) \\
+ C_2\kappa_2(q_{gbest} - q_i(t)) \\
+ C_3\kappa_3(q_{sgbest} - q_i(t)),
\end{split}
\label{eq:velocity}
\end{equation}

\begin{equation}
q_i(t) = q_i(t-1) + v_i(t),
\label{eq:update}
\end{equation}
where $W$ is the inertia weight used to control the impact of the previous velocity on the current velocity. $ C_1, C_2$, and $C_3$ indicate the learning factors, which balance the impact of different parts empirical on current velocity. $\kappa_1, \kappa_2$, and $\kappa_3$ are random values uniformly sampled from [0, 1], which are used to increase the randomness of the search.

Apart from the solution and velocity of a particle, the fitness function is crucial for optimizing in PSO algorithm. In this work, we adopt the following fitness function to evaluate each particle.

\begin{equation}
\min ~~ F(q) = {\rm Pr}_{\hat{y}}(A(x, q, M)),
\label{eq:fitness}
\end{equation}
where $A(x, q, M)$ denotes an applied function that paints the geometry with color $(red, green, blue)$ and transparency $\alpha$ at the coordinate of $o$ on the clean image $x$, where $M$ is a binary mask indicates the allowed modification area. ${\rm Pr}_{\hat{y}}()$ is the predicted label $\hat{y}$ probability of the target model $f$ on the input. By minimizing the $F(q)$, the confidence of prediction label $\hat{y}$ gradually decreases. We stop the search until it reaches the maximum iteration or finds the adversarial example. Algorithm \ref{alg1:rfa_attack} describes the optimization process.

\begin{algorithm}[t]
	\caption{Reflected Light Adversarial Attack (RFLA)}
	\label{alg1:rfa_attack}
	\textbf{Input}: input image $x$, target model $f$, max iteration $MaxIter$\\
	\textbf{Output}: Best solution $q^*$
	\begin{algorithmic}[1] 
		\STATE $q \leftarrow Initialization()$  ~~~$\triangleright$ Algorithm \ref{alg:gen_pop}
		\STATE $x_{adv} \leftarrow GenSample(q)$  ~~~$\triangleright$ Algorithm \ref{alg:gen_adv}
		\IF {$f(x_{adv})\neq f(x)$}
			\STATE return $q^*$ 
		\ENDIF 
		\FOR {$itr=i,...,MaxIter$}
			\STATE Update velocity vector $v$ by Equation \ref{eq:velocity}
			\STATE Update position vector $q$  by Equation \ref{eq:update}
    		\STATE $x_{adv} \leftarrow GenSample(q)$  ~~~$\triangleright$ Algorithm \ref{alg:gen_adv}
			\IF {$f(x_{adv})\neq f(x)$}
				\STATE return $q^*$
			\ENDIF 
		\ENDFOR
	\end{algorithmic}
\end{algorithm}

\subsection{Physical deployable attack} 

\begin{figure}[t]
	\centering
	\begin{minipage}{=.4\linewidth}
		\centering
		\includegraphics[width =1.\linewidth]{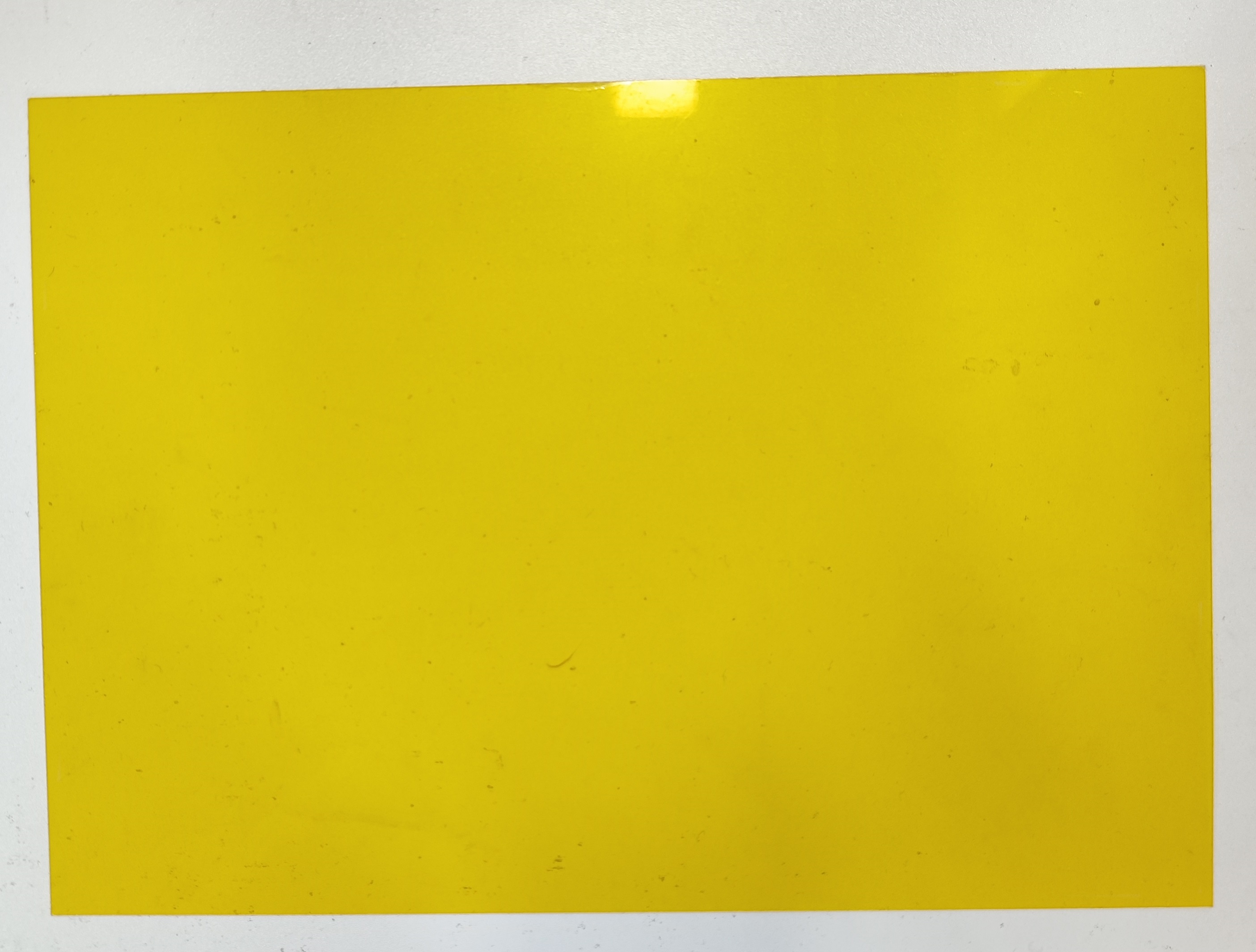}
		\centerline{\footnotesize (a)}
	\end{minipage}
	\begin{minipage}{=.4\linewidth}
		\centering
		\includegraphics[width =1.\linewidth]{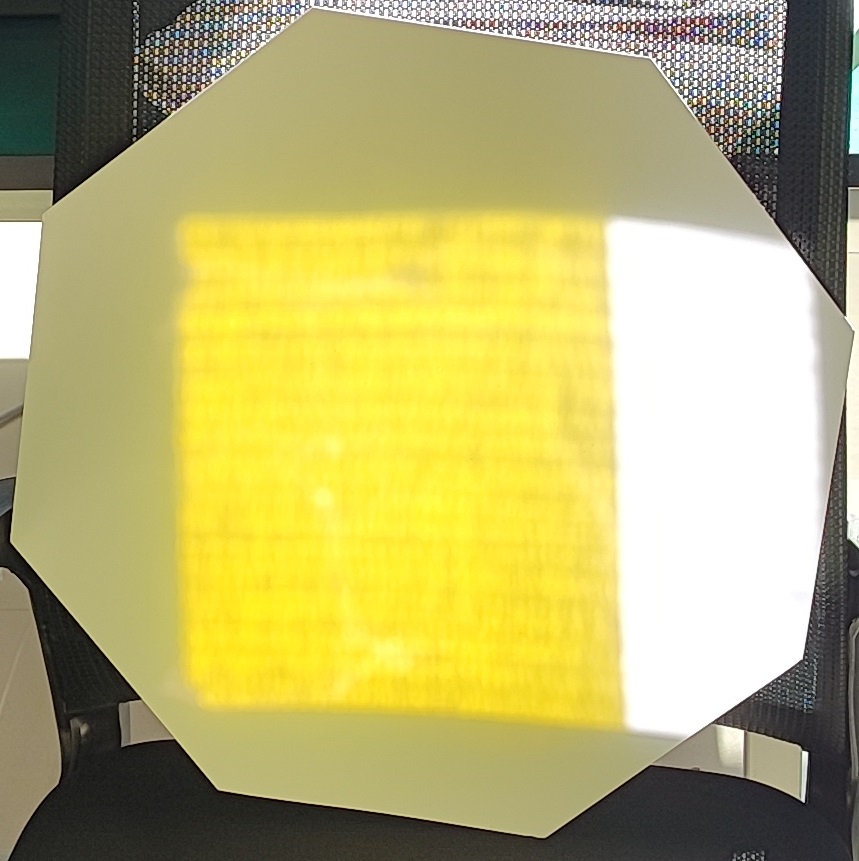}
		\centerline{\footnotesize (b)}
	\end{minipage}
\caption{The color discrepancy between the color transparency plastic sheet (a) and its reflected light(b).}
\label{fig:glasses}
\end{figure}

In digital worlds, we can construct $256^3$ color tuples by blending different RGB values, which is impractical in the physical world as the limitation of the device and material. Therefore, we constrain the search space of the color to ensure physically deployable. Specifically, we use seven color transparency plastic sheets to change the color of the reflected light. However, we find discrepancies exist between the color transparency plastic sheet and its reflected light (see Figure \ref{fig:glasses}), which may lead to attack failure. To decrease such discrepancy, we collect the light color reflected by the color transparency plastic sheet and adopt it as the searched color. In such a way, we can decrease color discrepancies when performing physical attacks.

\section{Experiments}

\subsection{Settings}
\textbf{Datasets:} To investigate the effectiveness of the proposed method, we conduct the digital attack on the ImageNet-compatible dataset provided by the NIPS 2017 adversarial competition\footnote{https://www.kaggle.com/c/nips-2017-non-targeted-adversarial-attack}, which includes 1000 images. Moreover, two commonly used traffic sign datasets: GTSRB \cite{gtsrb2011} and LISA \cite{lisa2012}, are also considered to investigate the extensibility of the proposed method.
 
\textbf{Target models:}  We conduct the proposed method on two tasks: image classification and traffic sign recognition. As for image classification, we select six ImageNet pre-trained networks: ResNet50 (RN50) \cite{resnet}, VGG16 \cite{vgg16}, DenseNet121 (DN121) \cite{densenet}, ResNeXt50 (RNX50) \cite{resnext29_16_64d}, WiderResNet50 (WRN50) \cite{Zagoruyko2016WRN} and SqueezeNet (SN) \cite{iandola2016squeezenet}, which are all provided by PyTorch \cite{paszke2019pytorch}. As for traffic sign recognition, we follow the setting reported in previous works \cite{eykholt2018robust,liu2019perceptual} to train GTSRB CNN and LISA CNN on GTSRB and LISA dataset, which obtains the accuracy of 95.06\% and 100\% on the test set, respectively.

\textbf{Evaluation metrics:} We adopt the attack success rate (ASR) as the evaluation metric, defined as the ratio of the number of the network's prediction flipped caused by adversarial examples to the total test dataset.

\textbf{Implementation details:} We adopt the OpenCV-Pyton package to plot different geometries in the clean image. For the settings of parameters of PSO, we set the max iteration number to 200, $C_1, C_2$, and $C_3$ set to 2.05, $W$ is set to 0.7298, the particle size and the geometry number at a circle are set to 50. The particle and velocity bound are provided in Appendix \ref{appendix:bound} (The upper bound of the transparency $\alpha$ is set to 0.7 to evade occluding the clean image.). Unless otherwise specified, the mask $M$ is set to all one value matrix in our experiments. All experiments were conducted on an NVIDIA RTX 3090ti 24GB GPU \footnote{Code will be available \url{https://github.com/winterwindwang/RFLA}.}.  

\subsection{Digital Adversarial Attacks}
In this section, we quantitatively and qualitatively evaluate the effectiveness of the proposed method in the digital world. For comparison, we adopt two patch-based attack methods: TPA \cite{yang2020patchattack} and DAPatch \cite{chen2022shape}; one line-based attack method Bezier \cite{giulivi2023adversarial}. TPA \cite{yang2020patchattack} utilized the feature texture image extracted from DNNs as the adversarial patch, which is pasted on the clean image, where the paste position is optimized by reinforcement learning. DAPatch \cite{chen2022shape} optimized the pattern and mask simultaneously, which can create a deformable shape adversarial patch. In contrast, Bezier \cite{giulivi2023adversarial} generated adversarial examples by scratching the bezier curve on the clean image, where the bezier curve depends on three points optimized by the optimizer (e.g., PSO). We reproduce the above three methods on the ImageNet-compatible dataset using the default settings.

\subsubsection{Quantitatively result}

\begin{table}[t]
\centering
\caption{Comparison results with the line-based method in terms of ASR (\%) on ImageNet-compatible dataset. The best results are highlighted with \textbf{bold}.}
\setlength\tabcolsep{2pt}
\begin{tabular}{ccccccc}
\hline
                                     & RN50 & VGG16 & DN121 & RNX50 & WRN50 & SN \\ \hline
Bezier \cite{giulivi2023adversarial} & 72 .4  & \textbf{77.7}    & 74.1    & 72.7    & 69.6    & \textbf{89.3} \\
RFLA-Line                            & \textbf{76.9} & 77.4  & \textbf{76.5}  & \textbf{75.7}  & \textbf{71.7}  &  89.2  \\ \hline
\end{tabular}
\label{tab:comp_line}
\end{table}

\begin{table}[t]
\centering
\caption{Comparison results with patch-based methods in terms of ASR (\%) on ImageNet-compatible dataset. The best results are highlighted with \textbf{bold}.}
\setlength\tabcolsep{2pt}
\small
\begin{tabular}{ccccccc}
\hline
                  & RN50 & VGG16 & DN121 & RNX50 & WRN50 & SN   \\ \hline
TPA \cite{yang2020patchattack}   & 66.1   & 36    & 40    & 24.7    & 23.4     & 44.6   \\
DAPatch \cite{chen2022shape}     & 74.3 & 71.6  & 79.3  & 73.7  & 76.7  & 56 \\
RFLA-Triangle  & 98.1    		& 97.8     		 & 97.2   			& 97.2    & 98.1      & 99.5     \\
RFLA-Rectangle & 99.3    		& 99.1           & 99.2   			& 98.9    & 99.1      & \textbf{99.8}     \\
RFLA-Pentagon  & \textbf{99.6}  & 99.1   		 & 99.2             & 99.2    & \textbf{99.4}  & \textbf{99.8}     \\
RFLA-Hexagon   & 99.5    		& \textbf{99.5}  & \textbf{99.5}    & \textbf{99.5}    &  \textbf{99.4}   & \textbf{99.8}     \\ \hline
\end{tabular}
\label{tab:comp_patch}
\end{table}

\begin{figure*}[htbp]
	\centering
	\begin{minipage}{=1.0\linewidth}
		\centering
		\includegraphics[width =1.\linewidth]{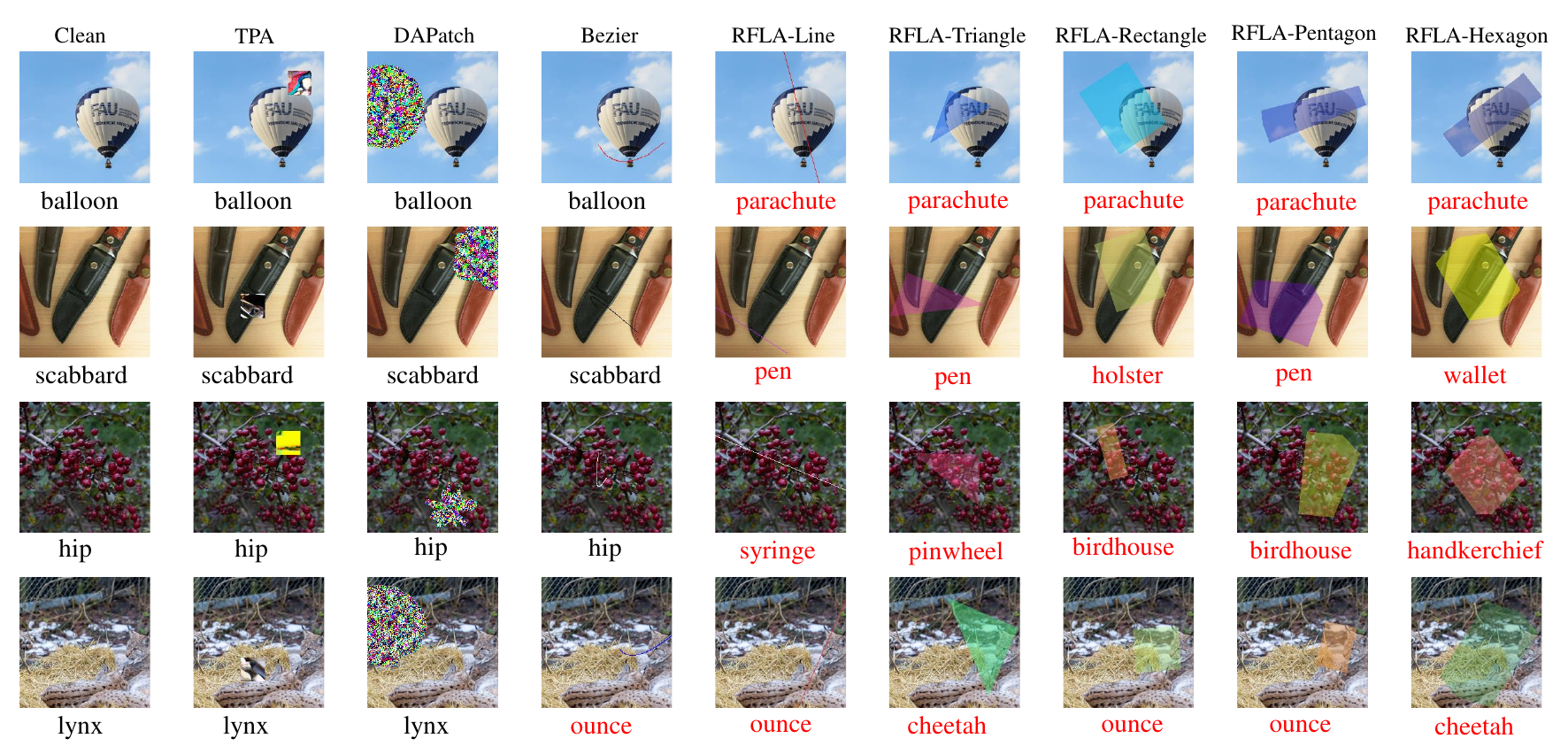}
	\end{minipage}
\caption{Visualization comparison of adversarial examples generated by different methods on ResNet50.}
\label{fig:visual_comparison}
\end{figure*}

Table \ref{tab:comp_line} reported the comparison results of the proposed RFLA-Line with the Bezier method. As we can observe, the proposed method outperforms the Bezier method on four of six models and obtains an improvement by 1.93\% in average ASR, indicating the proposed method's effectiveness. Additionally, we can get an 8.95\% improvement by widening the line thickness two times. Although the length of the Bezier curve may be shorter than ours, the discrepancy is trivial as modifications caused by the line is neglectable. Moreover, our method can be extended to more geometries.

Then, the comparison results of patch-based methods are listed in Table \ref{tab:comp_patch}. We conclude that the proposed geometry variants of RFLA outperform the existing method significantly. Specifically, the average ASR of the RFLA-triangle, RFLA-rectangle, RFLA-pentagon, and RFLA-hexagon are 97.98\%, 99.23\%, 99.38\%, and 99.53\%, obtaining the maximum improvement against TPA and DAPatch by 60.32\% and 27.52\%. We observe that the attack performance of comparison methods fails to achieve such results reported in their paper. The possible reason is that TPA may require two or more patches occluding 8\% of the image to achieve higher attack performance. As for DAPatch \cite{chen2022shape}, the position of the adversarial patch is ignored, which makes them fail to seek the model-decision-sensitive position. In contrast, our method simultaneously optimizes the position, geometry, and adversarial pattern, resulting in better performance. Moreover, the ASR gains with increasing vertex of geometry shape are attenuations, such as 20.08\% gains from Line (two vertexes) to Triangle(three vertexes), while only 0.15\% from Rectangle to Pentagon, which may be attributed to the improvement room of ASR is limited.

\begin{table}[t]
\setlength\tabcolsep{2pt}
\caption{Comparison results of transferability of adversarial examples generated by RNX50 in terms of ASR (\%) on ImageNet-compatible dataset. \textit{Item} indicates the white-box attack result, while the others are black-box results. The best results are highlighted with \textbf{bold}.}
\small
\begin{tabular}{ccccccc}
\hline
method          & RN50          & VGG16         & DN121         & RNX50                  & WRN50       & SN            \\ \hline
Bezier \cite{giulivi2023adversarial}  & \textbf{18.1} & \textbf{16.8} & 16.4          & \textit{72.7}          & 12.4        & 26.6          \\
RFLA-LINE       & 17.6          & 15.5          & \textbf{17.5} & \textit{\textbf{75.7}} & \textbf{14} & \textbf{27.1} \\ \hline
TPA \cite{yang2020patchattack}             & 28.9          & 36            & 26.7          & \textit{24.7}          & 20.4        & 45.4          \\
DAPatch \cite{chen2022shape} & 32.3          & \textbf{52.5} & 42.3          & \textit{73.7}          & 26.4        & 28.8          \\
RFLA-Triangle   & 37.8          & 34.9          & 33.1          & \textit{97.2}          & 33.7        & 50          \\
RFLA-Rectangle  & 47.6          & 43.2          & 44.3          & \textit{98.9}          & 45.52        & 59.9          \\
RFLA-Pentagon  & 47.5          & 45          & 46.4            & \textit{99.2} & 45.2        & 61.9            \\
RFLA-Hexagon   & \textbf{50.8} & 44            & \textbf{47.6} & \textit{\textbf{99.5}}          & \textbf{47} & \textbf{63.9} \\ \hline
\end{tabular}
\label{tab:transfer}
\end{table}

In addition, we compare the transferability of the proposed method with the comparison methods. Specifically, we use adversarial examples generated on RNX50 to attack other models. Evaluation results are reported in Table \ref{tab:transfer}. As we can observe, RFLA outperforms the comparison methods in most cases, and the magnitude of fall behind cases is small. Concretely, we obtain the maximum average improvement of ASR  of Bezier, TPA, and DAPatch are 0.28\% (RFLA-Line), 19.18\%, and 14.2\%, indicating the effectiveness of the proposed method. In addition, we provide other transferability comparison results in Appendix \ref{appendix:transfer}.

\subsubsection{Qualitatively result}
We provide the visualization result of adversarial examples generated by different methods in Figure \ref{fig:visual_comparison}. As we can observe, on the one hand, the Bezier and RFLA-Line obtain the most natural visuality quality, and the scratched line hardly observe at a glance. Meanwhile, RFLA-Line fools the DNNs in all displayed cases, but Bezier only has two success cases. On the other hand, TPA and DAPatch failed in all displayed cases. Such qualitative results can be used to explain their inferior attack performance. The adversarial patch generated by their method covers the noncontent areas, which may be insignificant to the model decision. Although the proposed method affects more image content than TPA and DAPatch, the covered contents are recognizable. In other words, our method does not modify the semantics of the image. We provide more visualization results of adversarial examples in Appendix \ref{fig:appendix_visual_comparison}.

\begin{figure}[t]
	\centering
	\begin{minipage}{=1.0\linewidth}
		\centering
		\includegraphics[width =1.\linewidth]{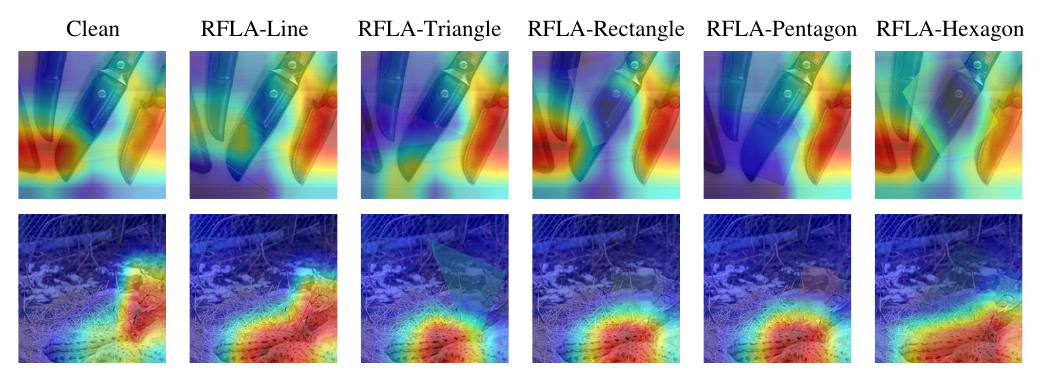}
	\end{minipage}
\caption{Model attention analysis of adversarial examples generated by RFLA on ResNet50.}
\label{fig:visual_cam}
\end{figure}

In addition, we use the Grad-CAM \cite{selvaraju2017grad} to investigate why the proposed method can work. Figure \ref{fig:visual_cam} illustrates the model attention visualization results. As we can observe, the painted geometry suppresses the model's attention areas, which makes the model output the wrong results. We also provide the visualization analysis result of comparison methods on model attention in Appendix \ref{fig:appendix_visual_cam}.

\subsection{Extend to Traffic Sign Recognition Task}
To further investigate the effectiveness of the proposed method, we use RFLA to attack the traffic sign recognition (TSR) model. Specifically, we collect 200 stop sign images from the GTSRB and LISA test set for evaluation. To avoid the geometry being out of the scope of the stop sign, we use a mask to indicate the allowable modification positions. We get the mask by averaging 200 test images and binary it. Table \ref{tab:attack_tsr} lists the digital adversarial performance. As we can observe, the proposed method obtains superior attack performances on two TSR models, especially for GTSRB CNN (100\% ASR). In addition, as for LISA-CNN, the attack performance increase with the geometries. One possible reason is that the affected area of the clean image is larger with the change of geometries under a similar circle. 

\begin{table}[t]
\centering
\caption{Quantitative results of RELA on TSR model in terms of ASR (\%).}
\begin{tabular}{ccc}
\hline
                  &  LISA-CNN & GTSRB-CNN   \\ \hline
RFLA-Triangle     &  68.5    & 100         \\
RFLA-Rectangle    &  92.5    & 100        \\
RFLA-Pentagon     &  93.5    & 100        \\
RFLA-Hexagon      &  97.5    & 100        \\  \hline
\end{tabular}
\label{tab:attack_tsr}
\end{table}

\begin{figure}[t]
	\centering
	\begin{minipage}{=1.0\linewidth}
		\centering
		\includegraphics[width =1.\linewidth]{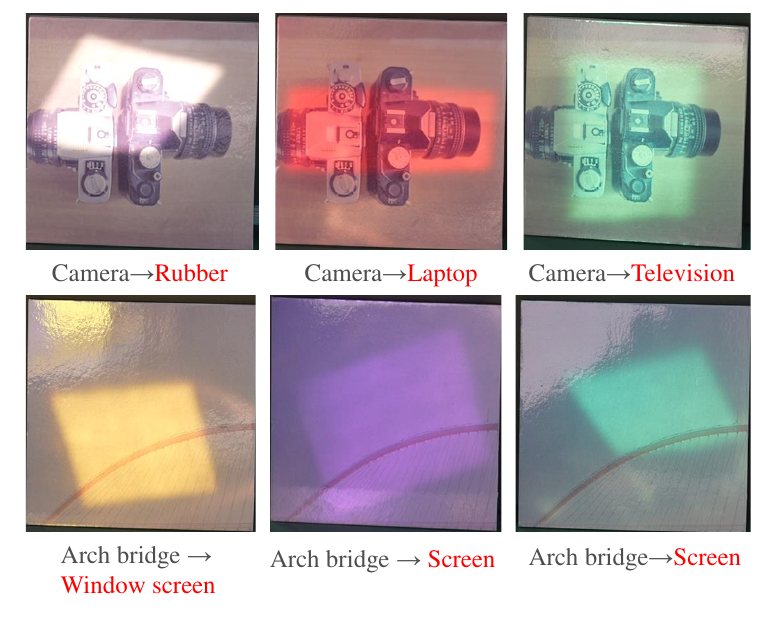}
	\end{minipage}
\caption{Physical adversarial examples.}
\label{fig:visual_physical}
\end{figure}

\begin{figure}[ht]
	\centering
	\begin{minipage}{=1.0\linewidth}
		\centering
		\includegraphics[width =1.\linewidth]{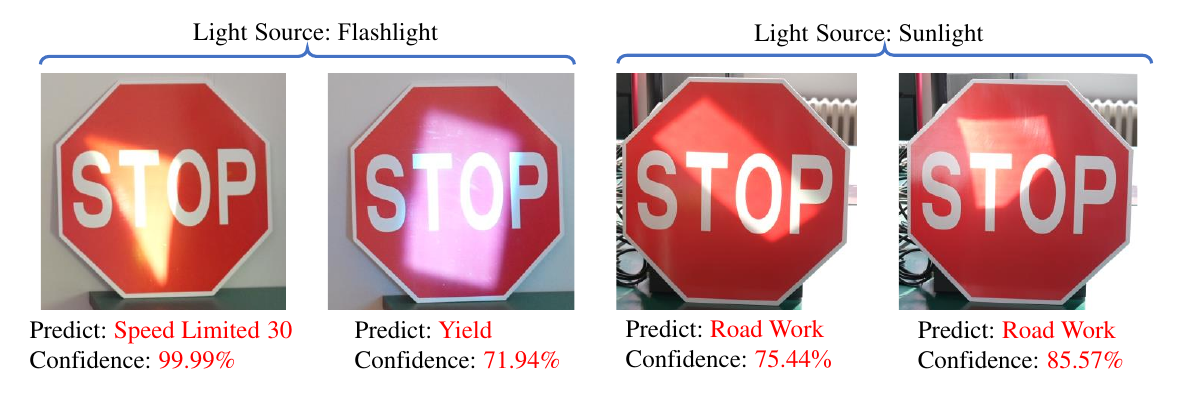}
	\end{minipage}
\caption{Examples of the "stop" sign with the reflected light and its prediction result on GTSRB CNN.}
\label{fig:visual_physical_tsr}
\end{figure}

\begin{table}[t]
\centering
\caption{Physical adversarial attacks under different light sources on different models.}
\begin{tabular}{cccc}
\hline
           & RN50   & VGG16     & DN121 \\ \hline
Sunlight   & 81.25\% & 81.25\%  & 81.25\%      \\
Flashlight & 87.5\%  & 87.5\%   & 87.5\%      \\ \hline
\end{tabular}
\label{tab:physical_result}
\end{table}
\subsection{Physical Adversarial Attacks}
Unlike the previous physical attacks that generate the adversarial pattern for the physically captured images, we generate the adversarial pattern (i.e., colored geometries) for digital images (the target model is RN50) and then reflect the light according to the optimized variables toward the corresponding printed images. In physical adversarial attacks, We use sunlight and a flashlight as the light source to mimic two different scenarios. We only evaluate one geometry (i.e., rectangle) for simplicity. Specifically, we randomly select six images from the dataset and generate the corresponding adversarial examples, where the color is fixed during the optimization for physical deployment. We use eight colors:  seven colors created by seven transparent color plastic sheets and one white reflected sunlight. Finally, we capture the physical adversarial examples from 2 meters away, collecting 48 images for each light source. 

Table \ref{tab:physical_result} lists the evaluation results. As we can observe, the ASR against the three models is above 80\% on physical adversarial examples created by two different light sources. Interestingly, we find that physical adversarial examples created by the reflected light against RN50 can consistently mislead the different models, indicating the reflected light is well-transferable even in the physical world. Figure \ref{fig:visual_physical} illustrates the physical adversarial examples. Furthermore, we study the effectiveness of reflected light attacks using sunlight and a flashlight on the TSR model. Figure \ref{fig:visual_physical_tsr} illustrates examples generated by different geometrical shapes.

\subsection{Ablation Study}

\textbf{Attack performance v.s. transparency.}
\begin{figure}[t]
	\centering
	\begin{minipage}{.45\linewidth}
		\centering
		\includegraphics[width =1.\linewidth]{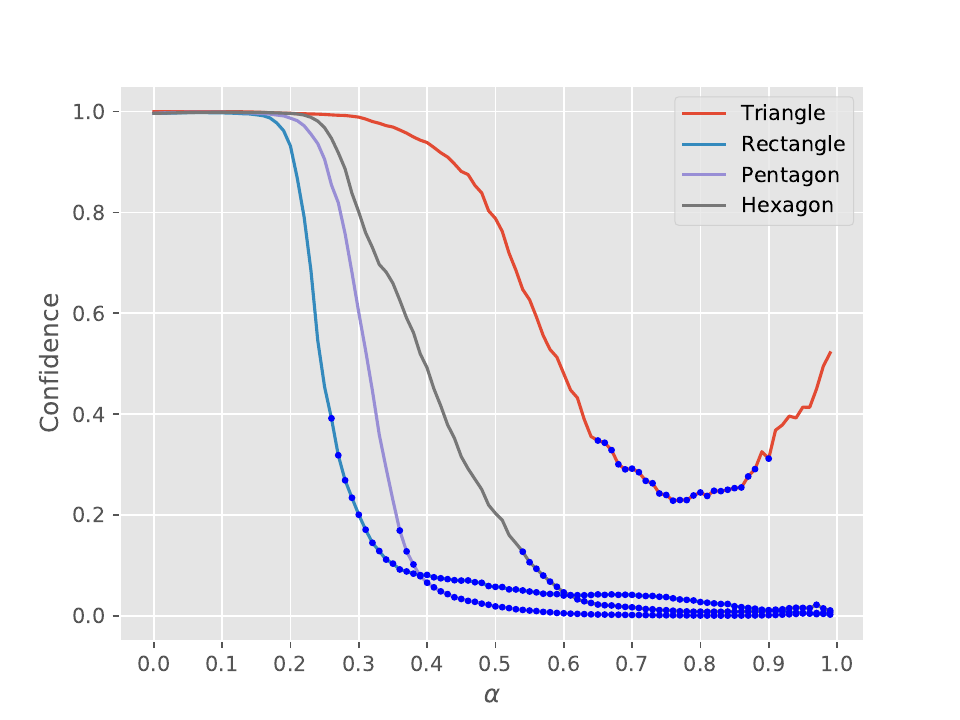}
		\centerline{\footnotesize (a) Confidence v.s. $\alpha$}
	\end{minipage}
	\begin{minipage}{.45\linewidth}
		\centering
		\includegraphics[width =1.\linewidth]{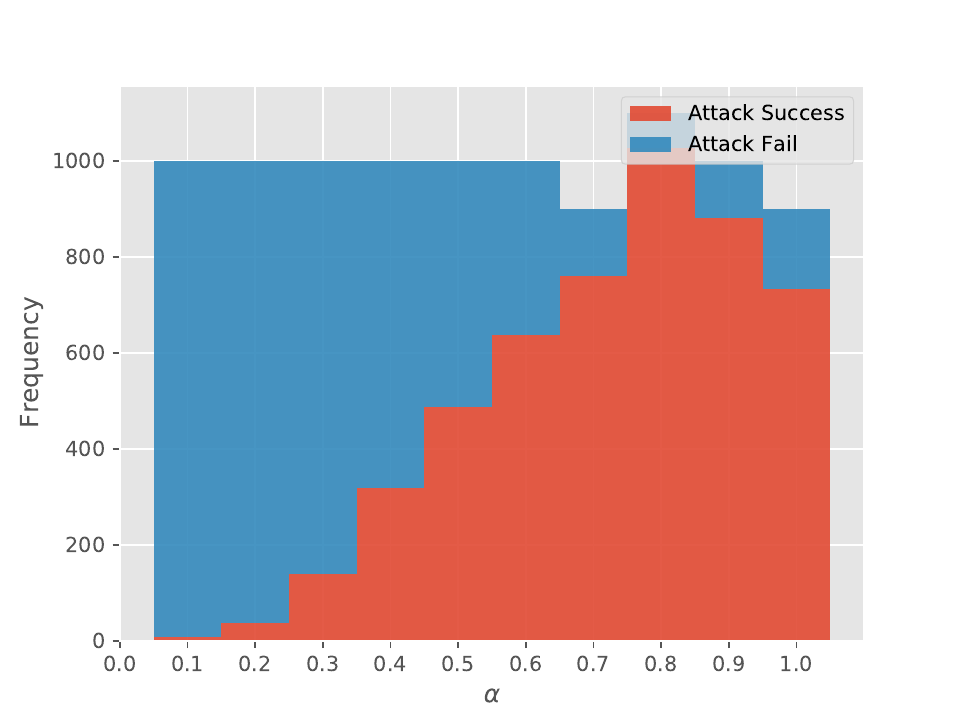}
		\centerline{\footnotesize (b) Frequency v.s. $\alpha$}
	\end{minipage}
\caption{From left to right: (a) The trend of ground-truth label confidence with changing transparency, where blue points denote the successful attack. (b) The frequency of successful attacks (RFLA-Triangle) with changing of transparency.}
\label{fig:ablation_alpha}
\end{figure}
The transparency $\alpha$ determined the cover intensity of the color. When $\alpha$ is set to one, the pixel value of the clean image at the specific position is substituted by the pure color, while the smaller value is the lower transparency of the color. We study how transparency changes the attack performance. Specifically, we fixed the other variables except for transparency, which traveled from [0,1] with a step size of 0.01. Figure \ref{fig:ablation_alpha} (a) illustrates the evaluation results of various geometries. As expectedly, the confidence of the ground-truth label decrease with enlarges of $\alpha$. In contrast, the attack performance (represented in the number of blue points) rises with the increases in transparency, as more content of the clean image is covered due to deeper pure color. Moreover, we statistics the frequency of successful and failed attacks of the RFLA-Triangle on 100 test images, which is depicted in Figure \ref{fig:ablation_alpha} (b), consistent with the previous analysis.

\textbf{Attack performance v.s. color.}
\begin{figure}[t]
	\centering
	\begin{minipage}{=.32\linewidth}
		\centering
		\includegraphics[width =1.\linewidth]{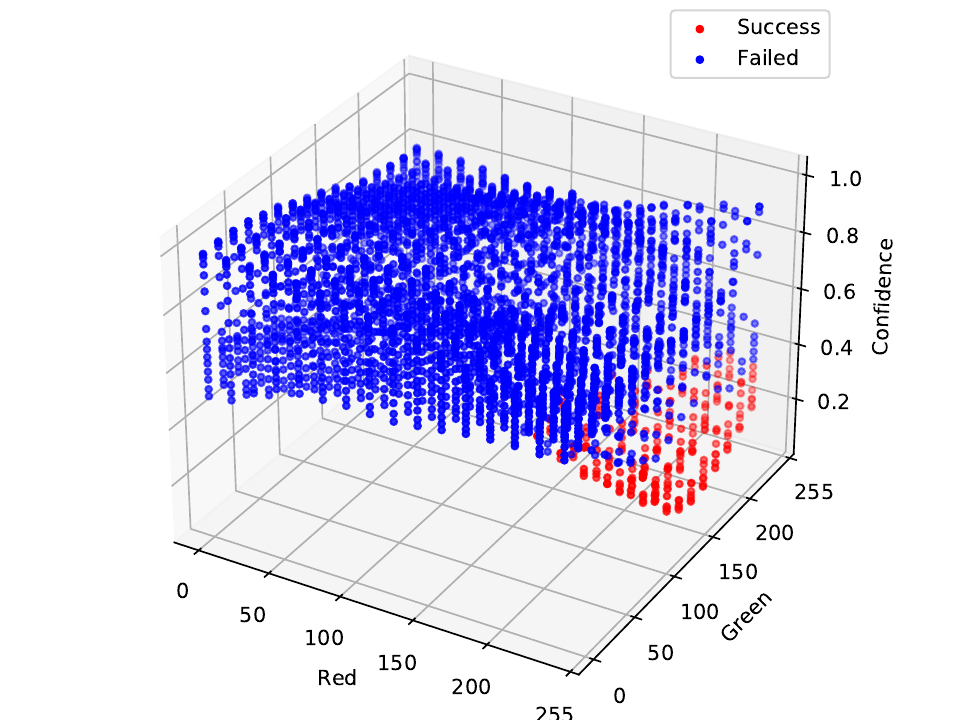}
	\end{minipage}
	\begin{minipage}{=.32\linewidth}
		\centering
		\includegraphics[width =1.\linewidth]{figures//Color_changed_imagenet_resnet50_rg}
	\end{minipage}
	\begin{minipage}{=.32\linewidth}
		\centering
		\includegraphics[width =1.\linewidth]{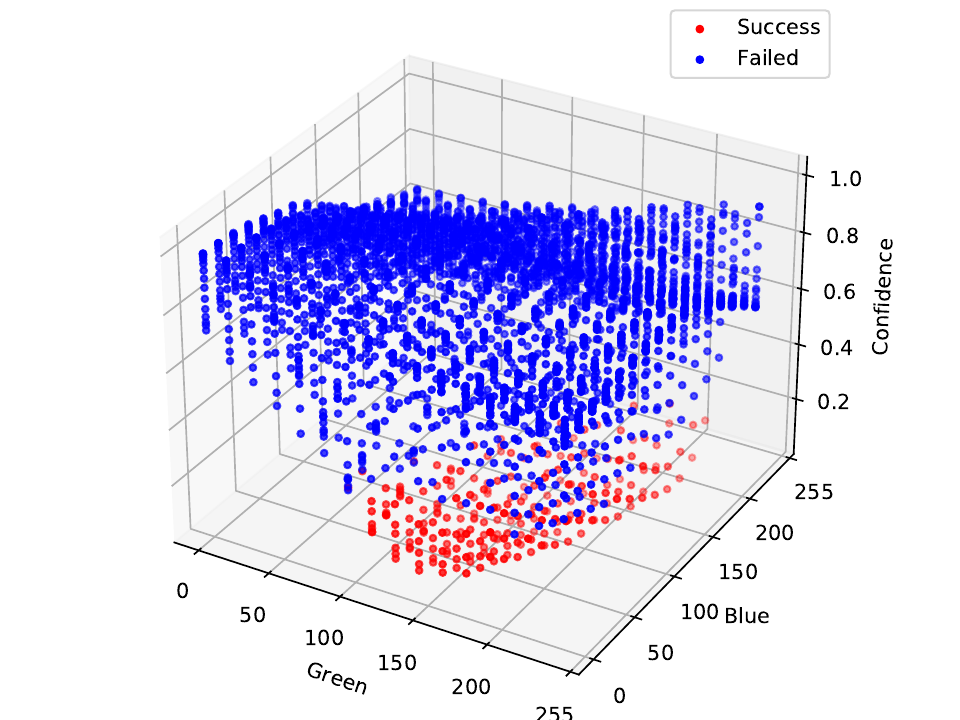}
	\end{minipage}
\caption{The confidence distribution of the ground-truth label on adversarial examples. Blue points denote the attack fail color tuple, while red points denote the attack success color tuple.}
\label{fig:ablation_color}
\end{figure}
The pattern of adversarial perturbation is crucial for a successful attack. Unlike the previous works that optimize the pixel-wise perturbation, we focus on the channel-wise perturbation (perturb a channel with one value) as we must ensure the perturbation is physically realizable by reflecting the light. Furthermore, channel-wise perturbation is visually more acceptable than pixel-wise perturbation. Specifically, we select the color tuple set in intervals of 16 pixels across the three RGB channels to investigate how color influences the attack performance. Figure \ref{fig:ablation_color} illustrates the evaluation results. As we can observe, the success cases almost cluster in specific areas near the searched optimal color tuple when other variables are fixed. In other words, the optimal color tuple has certain robustness to a slight change of color, which makes our attacks can undertake some distortions when applied in the physical world.

\textbf{Attack performance v.s. position.}
\begin{figure}[t]
	\centering
	\begin{minipage}{=.25\linewidth}
		\centering
		\includegraphics[width =1.\linewidth]{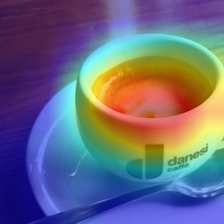}
		\centerline{\footnotesize (a)}
	\end{minipage}
	\begin{minipage}{=.25\linewidth}
		\centering
		\includegraphics[width =1.\linewidth]{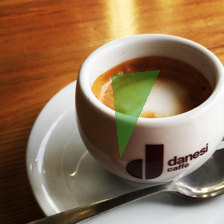}
		\centerline{\footnotesize (b)}
	\end{minipage}
	\begin{minipage}{=.35\linewidth}
		\centering
		\includegraphics[width =1.\linewidth]{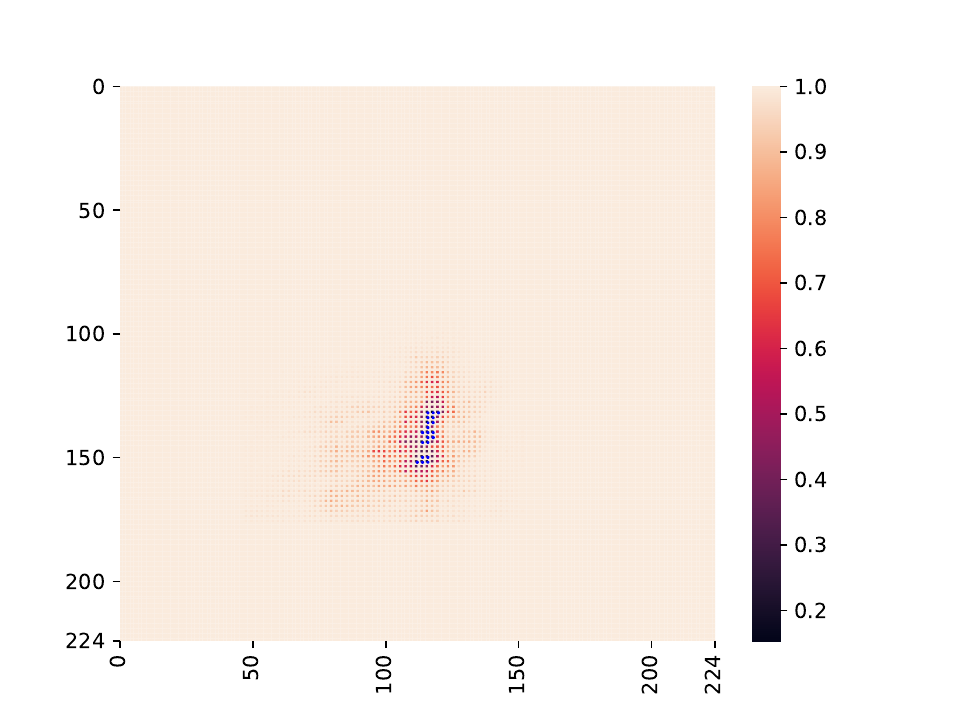}
		\centerline{\footnotesize (c)}
	\end{minipage}
\caption{From left to right: (a) Model attention on the clean image; (b) Adversarial examples; (c) Prediction confidence distribution of the clean image with the change of position, where blue points (the center of the circle) denote the successful attacks.}
\label{fig:ablation_position}
\end{figure}
To investigate the influence of the position of the adversarial perturbation to attack performance, we fixed the optimal variable except for position. Then, we sample the position in intervals of two steps. Furthermore, we also give the Grad-Cam for comparison. Figure \ref{fig:ablation_position} provides the evaluation results. As we can see, the adversarial geometry plotted around the content areas significantly drops the prediction confidence of the model on the clean image. Meanwhile, the attack success area is consistent with the model attention area, which indicates that our method can automatically locate the model attention areas to perform attacks.

\section{Conclusion}
In this paper, we propose a novel reflected light attack to realize effective and stealthy attacks in both digital and physical worlds,  which may impose potential risks to automatic driving systems. Specifically, to explore how to control the reflected light's position, geometry, and pattern, we exploit the characteristic of the circle and propose a general framework based on the circle. To create a geometry, we first generate a specific number of angles to construct the point in circumference, followed by applying point symmetry against the center of a circle to generate a new point. These obtained points fence a geometrical shape where the fill color and transparency are optimized. Finally, we apply the PSO algorithm to find the best position, geometry, fill color, and transparency. Experiment results on digital and physical attacks verify the effectiveness of the proposed method. Moreover, our method can not only use sunlight but also can use flashlights to perform physical attacks for adapting to different environments. 

\textbf{Limitations.} Though the reflected-light attack can perform in different environments, it is hard to remain effective in bad weather, such as fog and rain. A more penetrating light source (e.g., the traffic light and foglight) may work in such conditions.

\textbf{Potential negative societal impact and mitigation.} Similar to other types of attack, the adversarial attack is inevitable to cause potential security risks, especially for those physically deployed systems. However, we aim to arouse people's attention to such related applications and then encourage people to develop defense techniques to counter the reflected-light attack. To thwart the RFLA attack proposed in this paper, one can develop multimodal-based DNN systems.

\section{Acknowledgments}
The authors are grateful to the anonymous reviewers for their insightful comments. This work was supported by the National Natural Science Foundation of China (No.11725211 and 52005505).

{\small
\bibliographystyle{ieee_fullname}
\bibliography{references}
}

\clearpage
\appendix

\renewcommand\thefigure{\Alph{section}\arabic{figure}}
\renewcommand\thealgorithm{\Alph{section}\arabic{algorithm}}
\renewcommand\thetable{\Alph{section}\arabic{table}}
\section{Implementation details} \label{appendix:bound}

In this section, we first introduce the lower and upper bounds of the particle and velocity. Recall that a particle $q$ represents an optimization variable tuple $(x, y, r, \alpha, red, green, blue, a_1)$, where the lower bound is (0, 0, 10, 0, 0, 0, 0, 0, 0, 0) and the upper bound is (H/2, W/2, 0.4$\times$min(W, H), 0.7, 255, 255, 255, 360) except for the line. As for the RFLA-Line, the transparency $\alpha$ is set to 1. The reason is that the modification of the clean image caused by the RFLA-Line geometry shape merely has a few numbers of pixels, even if the original pixels are replaced. Velocity controls the movement speed of particles: a large velocity speed would early lead the particle to reach the bound, while a small velocity makes the particle move slowly, requiring more optimizing time. Thus, we set them in terms of the concrete meaning of the variable. Specifically, we set the upper bound of velocity as follows: coordination and radius of the circle are set to 5, 5, and 10; the transparency $\alpha$ is set to 0.05; the color is set to 5; the angle is set to 10, i.e., the initialization of velocity's upper bound is set to $v_{upper} = (5, 5, 10, 0.05, 5, 5, 5, 10)$. In contrast, the lower bound is set to $v_{lower}=-v_{upper}$.

The initialization of the proposed method is significant to optimization, which describes the variable that requires to be optimized. Algorithm \ref{alg:gen_pop} describes the initialization of particles and the corresponding velocity generation for different geometries. Specifically, we generate population size $S$ particles, i.e., the center $o(x,y)$ and radius $r$ of the circle. For each specific circle, we generate subpopulation size $S_{sub}$ geometries, which consists of $\alpha, red, green, blue, a_1$. Note that we fixed the coordinate and radius when spawning different topologies of the same geometrical shape at a specific circle. Therefore, we devise a novel variable to record which circle can generate the optimal geometrical shapes from an overall viewpoint, i.e., the $q_{sgbest}$, which is defined as 

\begin{equation}
q_{sgbest} = \arg \min_{i \in S} \sum^{S_{sub}}_{j=1} F(q_{i,j}).
\label{eq:q_best}
\end{equation}

Moreover, the defination of the $q_{gbest}$ and $q_{pbest}$ is expressed as follow

\begin{equation}
q_{gbest} = \arg \min_{i \in S, j \in S_{sub}} F(q_{i,j}).
\end{equation}

\begin{equation}
q_{pbest}^i = \arg \min_{j \in S_{sub}} F(q_{i,j}).
\end{equation}

After generating the particle, we use the Algorithm \ref{alg:gen_adv} to generate adversarial examples. Specifically, we first obtain the point on the circle by cosine and sine function with the coordinate and radius and an angle, then calculate its symmetry point with respect to the center points. Repets until generating enough points that the geometry shape required. Then, we sort the point set to avoid engendering the intercross edge. Finally, we use the OpenCV package to plot the geometry on the clean image. 

\begin{algorithm}[ht]
	\caption{Population Initialization}
	\label{alg:gen_pop}
	\textbf{Input}: mask $M$, population size $S$, sub population size $S_{sub}$ \\
	\textbf{Output}: $q_{i,j}, v_{i,j}, ~i \in [1,...,S], j \in [1,...,S_{sub}]$
	\begin{algorithmic}[1] 
		\STATE  $ q \leftarrow$ []
		\STATE  $ v \leftarrow$ []
		\FOR {$i=1,...,S$}
		    \STATE Randomly sample a radius $r$ in $[r_{lower}, r_{upper}]$
			\STATE Sample a center $(x, y)$ of the circle by $[r, x_{upper}-r]$ and $[r, y_{upper}-r]$
			\WHILE{$M(x,y)==0$}
			\STATE Resample a center $(x, y)$ of the circle
			\ENDWHILE
			\STATE Random initialize $v_{x},v_{y},v_{r}$ from $[V_{lower}, V_{upper}]$
			\FOR {$j=0,...,S_{sub}$}
				\STATE Initialize a transparency $\alpha$, a angle $a_1$ and fill color $(red,green,blue)$
				\STATE Initialize velocity $v_{\alpha}, v_{a_1}, v_{red}, v_{green}, v_{blue}$
				\IF {shape == line}
					 \STATE  $q_{i,j} \leftarrow$ $(x,y,r, \alpha, red,green,blue, a_1)$
					 \STATE $v_{i,j}  \leftarrow$ $(v_x,v_y,v_r, v_{\alpha}, v_{red}, v_{green}, v_{blue}, v_{a_1})$
				\ENDIF
				\IF {shape == triangle \OR shape == rectangle}
					 \STATE Initialize a new angle $a_1$
					 \STATE Initialize a new angle velocity $v_{a_2}$					 
					 \STATE $q_{i,j} \leftarrow$ $(x,y,r, \alpha, red,green,blue a_1, a_2)$
					 \STATE $v_{i,j}  \leftarrow$ $(v_x,v_y,v_r, v_{\alpha}, v_{red}, v_{green}, v_{blue}, v_{a_1}, v_{a_2})$					 
				\ENDIF
				\IF {shape == pentagon \OR shape == hexagon}
					 \STATE Initialize a new angle $a_1$ and $a_2$
					 \STATE Initialize a new angle velocity $v_{a_2}, v_{a_3}$	
					 \STATE $q_{i,j} \leftarrow$ $(x,y,r, \alpha, red,green,blue, a_1, a_2, a_3)$
					 \STATE $v_{i,j}  \leftarrow$ $(v_x,v_y,v_r, v_{\alpha}, v_{red}, v_{green}, v_{blue}, v_{a_1}, v_{a_2}, v_{a_3})$					 
				\ENDIF
		\ENDFOR
	\ENDFOR
	\end{algorithmic}
\end{algorithm}

\begin{algorithm}[t]
	\caption{Generate Adversarial Examples}
		\label{alg:gen_adv}
	\textbf{Input}: Populations $q$ \\
	\textbf{Output}: $x_{adv}$
	\begin{algorithmic}[1] 
		\STATE $x_{adv} \leftarrow$ []
		\FOR {$i=1,...,S$}
			\FOR {$j=1,...,S_{sub}$}
				\IF {shape == line}
					 \STATE $(x, y, radius, \alpha, r, g, b, a_1) \leftarrow q_{i,j}$
					 \STATE Get the point set ${p_1, p'_1}$
				\ENDIF
				\IF {shape == triangle}
					 \STATE $(x, y, radius, \alpha, r, g, b, a_1, a_2) \leftarrow q_{i,j}$
					 \STATE Get the point set ${p_1, p_2, p'_1}$
				\ENDIF
				\IF {shape == triangle}
					 \STATE $(x, y, radius, \alpha, r, g, b, a_1, a_2) \leftarrow q_{i,j}$
					 \STATE Get the point set ${p_1, p_2, p'_1, p'_2}$
				\ENDIF
				\IF {shape == pentagon}
					 \STATE $(x, y, radius, \alpha, r, g, b, a_1, a_2, a_2) \leftarrow q_{i,j}$
					 \STATE Get the point set ${p_1, p_2, p_3, p'_1, p'_3}$
				\ENDIF
				\IF {shape == hexagon}
					 \STATE $(x, y, radius, \alpha, r, g, b, a_1, a_2, a_2) \leftarrow q_{i,j}$
					 \STATE Get the point set ${p_1, p_2, p_3, p'_1, p'_2, p'_3}$
				\ENDIF
				\STATE Sorted the point set via the Minimum Points Distance algorithm
				\STATE $x_{adv} \leftarrow $ call function \textit{fillpoly} of OpenCV-Package with the input $x$ and particle $q$.
		\ENDFOR
	\ENDFOR
	\end{algorithmic}
\end{algorithm}

\section{Experiment and result analysis} \label{appendix:extra_exp}

To demonstrate the effectiveness of the proposed algorithm, we conducted an ablation study with random search algorithm. We fixed the light color (i.e., set to white sunlight), and only searched the optimal position and geometry.  Table \ref{tab:randomly_baseline} reports the evaluation results, which suggests that our method can significantly outperform the random baseline, where our achieves the average ASR of 95.13\%, obtaining 13.63\% higher than the random baseline.
\begin{table}[h]
\setlength\tabcolsep{2pt}
\centering
\footnotesize
\caption{Comparison results of random baselines and ours.}
\label{tab:randomly_baseline}
\begin{tabular}{ccccccc}
\hline
              & RN50 & VGG16 & DN121 & RNX50 & WRN50 & SN   \\ \hline 
Random-Tri-White & 79.2\%     & 79.2\%  & 82.5\%        & 78.2\%      & 79.2\%  & 90.7\%    \\ 
RFLA-Tri-White & 95.4\%     & 94.8\%  & 94.6\%        & 93\%        & 95.1\%  & 97.9\%           \\ \hline
\end{tabular}
\end{table}

We also attempt to implement the targeted attack by modifying the fitness function. Specifically, we perform the experiment on WRN50 and set the targeted label by $(y+1)\%1000$. Table \ref{tab:targeted_attack} lists the targeted attack results. As we can observe that the targeted attack is not satisfying comparing with the white-box attack, where the average ASR is
21.26\%. We speculate the reason may attribute to the pure color perturbation is hard to implement the targeted attack. Moreover, we also found that the optimization time of the black-box attack is enormously increased compared with the white-box attack.

\begin{table}[h]
\centering
\caption{Evaluation results of target attack on WRN50.}
\begin{tabular}{ccccc}
\hline
           &Triangle & Rectangle & Pentagon & Hexagon \\ \hline
WRN50	   & 17.8\%   & 21.9\%    & 22.7\%   & 22.64\%  \\ \hline
\end{tabular}
\label{tab:targeted_attack}
\end{table}

\begin{figure*}[ht]
	\centering
	\begin{minipage}{=1.0\linewidth}
		\centering
		\includegraphics[width =1.\linewidth]{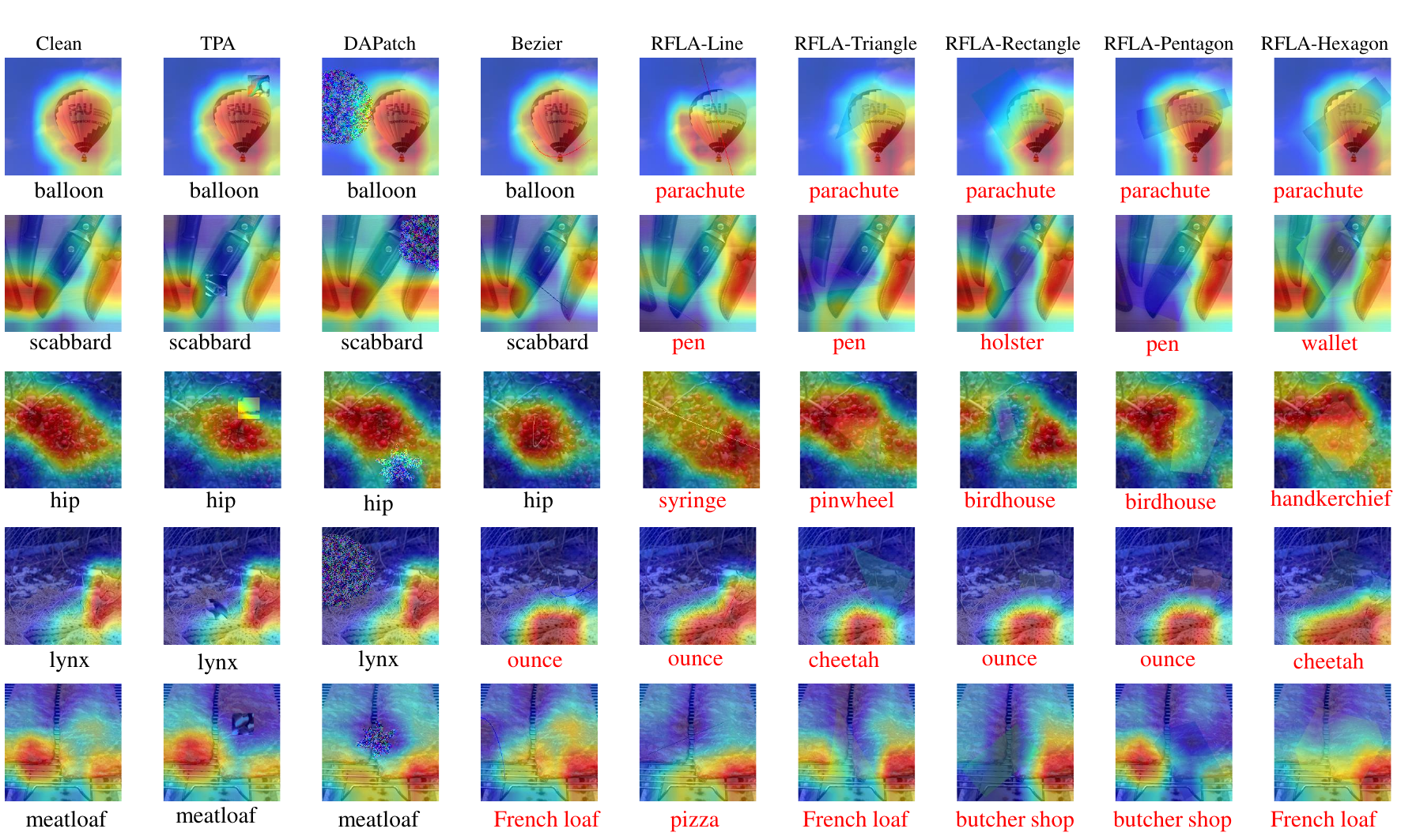}
	\end{minipage}
\caption{Model attention analysis of adversarial examples generated by different methods on ResNet50.}
\label{fig:appendix_visual_cam}
\end{figure*}

Model attention analysis can reveal how the attack algorithm works. To provide a more complete visualization comparison results of different attack methods. We use the Grad-CAM tool to analyze changes in the class activation map caused by different attack algorithms. Specifically, we focus on the CAM of the original prediction class since the predicted class of the adversarial examples is different from the original prediction class. Thus, the CAM is also different. In contrast,  the changes in the CAM of the original prediction can reflect the attack function. The comparison result is illustrated in Figure \ref{fig:appendix_visual_cam}. As we can observe, the proposed method can disperse the CAM while the other method can not. Take a careful look at the CAM of the proposed method, the plotted geometry suppresses the original CAM, where the region is the region with semantic content. Therefore, the proposed method can obtain superior attack performance.

In addition, we provide the complete transferability comparison result in Table \ref{appendix:transfer}. As we can see, the proposed method achieves the best average ASR on both white-box and black-box attacks. One possible reason for the better transferability of the proposed method is that the proposed method can automatically locate the region of the model decision that is common for different models. Moreover, the proposed method is different from the full-pixel imperceptible perturbation, we generate adversarial examples by modifying partial image regions with the transparency color. Therefore, the image content of the modified region by the proposed method is maintained, which is the main difference compared to the patch-based attack (e.g., TPA and DAPatch).

\begin{figure*}[ht]
	\centering
	\begin{minipage}{=1.0\linewidth}
		\centering
		\includegraphics[width =1.\linewidth]{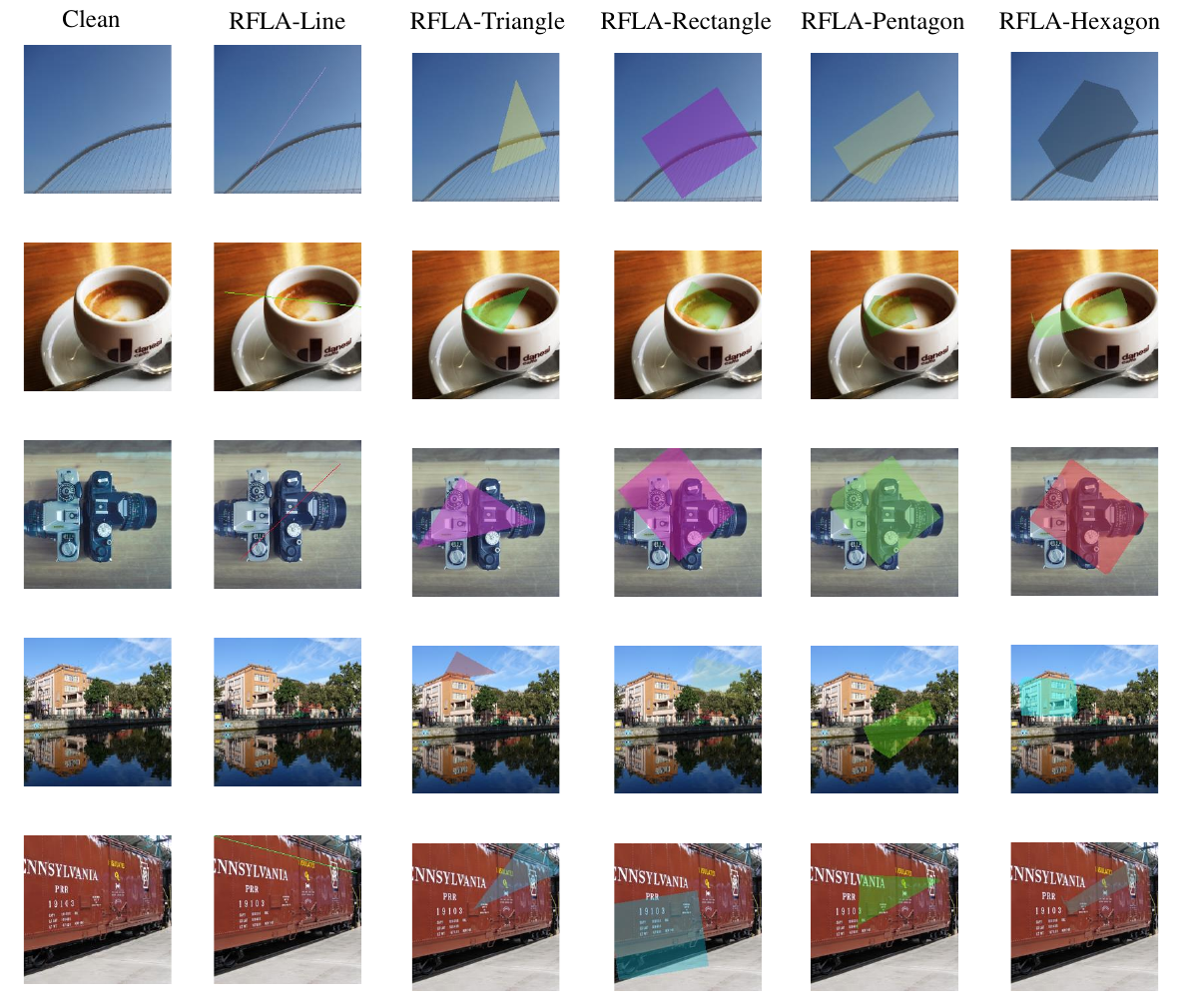}
	\end{minipage}
\caption{Visualization comparison of adversarial examples generated by RFLA on ResNet50.}
\label{fig:appendix_visual_comparison}
\end{figure*}

\begin{table*}[]
\centering
\caption{Comparison results of transferability of adversarial examples generated by different methods in terms of ASR (\%) on ImageNet-compatible dataset. \textit{Item} indicates the white-box attack result, while the others are black-box results. The best results are highlighted with \textbf{bold}.}
\label{appendix:transfer}
\begin{tabular}{cclllllll}
\hline
                       & method         & \multicolumn{1}{c}{RN50} & \multicolumn{1}{c}{VGG16} & \multicolumn{1}{c}{DN121} & \multicolumn{1}{c}{RNX50} & \multicolumn{1}{c}{WRN50} & \multicolumn{1}{c}{SN} & \multicolumn{1}{c}{AVG} \\ \hline
\multirow{8}{*}{RN50}  & TPA            & \textit{66.1}            & \textbf{50.4}             & 42.7                      & \textbf{41.6}             & 38.6                      & 55                     & 45.66                   \\
                       & DAPatch        & \textit{74.3}            & 47.4                      & 44.5                      & 48                        & 31.9                      & 58.6                   & 46.08                   \\
                       & Bezier         & \textit{72.4}            & 15.8                      & 15.5                      & 17.8                      & 13.8                      & 27.4                   & 18.06                   \\
                       & RFLA-LINE      & \textit{76.9}            & 14.5                      & 15.5                      & 14.1                      & 14.1                      & 27.7                   & 17.18                   \\
                       & RFLA-Triangle  & \textit{98.1}            & 32.2                      & 33                        & 29.4                      & 30.6                      & 48.5                   & 34.74                   \\
                       & RFLA-Rectangle & \textit{99.3}            & 43.8                      & 44.8                      & 37.5                      & 40                        & 60                     & 45.22                   \\
                       & RFLA-Pengtagon & \textit{\textbf{99.6}}   & 42.4                      & 42.7                      & 35.2                      & 38.5                      & \textbf{62.8}          & 44.32                   \\
                       & RFLA-Hextagon  & \textit{99.5}            & 43.7                      & \textbf{46.9}             & 38.1                      & \textbf{44.2}             & 62.4                   & \textbf{47.06}          \\ \hline
\multirow{8}{*}{VGG16} & TPA            & 30                       & \textit{36}               & 29                        & 26                        & 22                        & 46                     & 30.6                    \\
                       & DAPatch        & 24.9                     & \textit{71.6}             & 35                        & 31.1                      & 19.8                      & \textbf{76.4}          & 37.44                   \\
                       & Bezier         & 12                       & \textit{77.7}             & 12.8                      & 12.9                      & 9.3                       & 29.2                   & 15.24                   \\
                       & RFLA-LINE      & 14.4                     & \textit{77.4}             & 15.9                      & 14.1                      & 11.6                      & 30                     & 17.2                    \\
                       & RFLA-Triangle  & 28.7                     & \textit{97.8}             & 26.9                      & 23                        & 23.2                      & 47.8                   & 29.92                   \\
                       & RFLA-Rectangle & 35.1                     & \textit{99.1}             & 30.1                      & 28.1                      & 30.4                      & 52.3                   & 35.2                    \\
                       & RFLA-Pengtagon & 41.4                     & \textit{99.1}             & 40.7                      & 33.1                      & 36.7                      & 61.5                   & 42.68                   \\
                       & RFLA-Hextagon  & \textbf{43.2}            & \textit{\textbf{99.5}}    & \textbf{43.1}             & \textbf{37.1}             & \textbf{39}               & 65.4                   & \textbf{45.56}          \\ \hline
\multirow{8}{*}{DN121} & TPA            & 33.4                     & 40.2                      & \textit{40}               & 27.6                      & 25.6                      & 48.8                   & 35.12                   \\
                       & DAPatch        & 28.8                     & \textbf{50.7}             & \textit{79.3}             & 37.2                      & 21.1                      & 74.4                   & 42.44                   \\
                       & Bezier         & 16.6                     & 14.4                      & \textit{74.1}             & 14.1                      & 10.6                      & 27.5                   & 16.64                   \\
                       & RFLA-LINE      & 15.4                     & 14.9                      & \textit{76.5}             & 15.7                      & 11.7                      & 26.9                   & 16.92                   \\
                       & RFLA-Triangle  & 31.7                     & 29.8                      & \textit{97.2}             & 25.5                      & 26.8                      & 46.9                   & 32.14                   \\
                       & RFLA-Rectangle & 42.7                     & 41.6                      & \textit{99.2}             & 35                        & 36.6                      & 62.3                   & 43.64                   \\
                       & RFLA-Pengtagon & 45.2                     & 43.1                      & \textit{99.2}             & 34.8                      & 38.5                      & 60.3                   & 44.38                   \\
                       & RFLA-Hextagon  & \textbf{45.2}            & 43.8                      & \textit{\textbf{99.5}}    & \textbf{38}               & \textbf{39.6}             & \textbf{63.3}          & \textbf{45.98}          \\ \hline
\multirow{8}{*}{RNX50} & TPA            & 28.9                     & 36                        & 26.7                      & \textit{25}               & 20.4                      & 45.4                   & 31.48                   \\
                       & DAPatch        & 32.3                     & \textbf{52.5}             & 42.3                      & \textit{73.7}             & 26.4                      & 28.8                   & 36.46                   \\
                       & Bezier         & 18.1                     & 16.8                      & 16.4                      & \textit{72.7}             & 12.4                      & 26.6                   & 18.06                   \\
                       & RFLA-LINE      & 17.6                     & 15.5                      & 17.5                      & \textit{75.7}             & 14                        & 27.1                   & 18.34                   \\
                       & RFLA-Triangle  & 37.8                     & 34.9                      & 33.1                      & \textit{97.2}             & 33.7                      & 50                     & 37.9                    \\
                       & RFLA-Rectangle & 47.6                     & 43.2                      & 44.3                      & \textit{98.9}             & 45.2                      & 59.9                   & 48.04                   \\
                       & RFLA-Pengtagon & 47.5                     & 45                        & 46.4                      & \textit{99.2}             & 45.2                      & 61.9                   & 49.2                    \\
                       & RFLA-Hextagon  & \textbf{50.8}            & 44                        & \textbf{47.6}             & \textit{\textbf{99.5}}    & \textbf{47}               & \textbf{63.9}          & \textbf{50.66}          \\ \hline
\multirow{8}{*}{WRN50} & TPA            & 31.4                     & 36.6                      & 27.3                      & 24.7                      & \textit{23.4}             & 44                     & 32.8                    \\
                       & DAPatch        & 40.2                     & \textbf{48.4}             & \textbf{49.3}             & \textbf{52.2}             & \textit{76.7}             & 60.8                   & 50.18                   \\
                       & Bezier         & 19.4                     & 17.7                      & 14.6                      & 18.7                      & \textit{69.6}             & 25.1                   & 19.1                    \\
                       & RFLA-LINE      & 17.8                     & 15                        & 16.8                      & 16.2                      & \textit{71.7}             & 25.2                   & 18.2                    \\
                       & RFLA-Triangle  & 36.1                     & 31.7                      & 35.3                      & 31.7                      & \textit{98.1}             & 47.4                   & 36.44                   \\
                       & RFLA-Rectangle & 45.3                     & 41.1                      & 43.3                      & 38.8                      & \textit{99.1}             & 58.9                   & 45.48                   \\
                       & RFLA-Pengtagon & 48.9                     & 41.8                      & 46.8                      & 40                        & \textit{99.4}             & 61.4                   & 47.78                   \\
                       & RFLA-Hextagon  & \textbf{51}              & 47.3                      & 48.1                      & 40.6                      & \textit{99.4}             & \textbf{64.7}          & \textbf{50.34}          \\ \hline
\multirow{8}{*}{SN}    & TPA            & \textbf{29.8}            & 28.5                      & 29.7                      & \textbf{23}               & 22.5                      & \textit{44.6}          & 26.7                    \\
                       & DAPatch        & 17.3                     & \textbf{30.1}             & 23.8                      & 22.2                      & 14.3                      & \textit{56}            & 21.54                   \\
                       & Bezier         & 10.3                     & 12.2                      & 11.3                      & 9.1                       & 8                         & \textit{89.3}          & 10.18                   \\
                       & RFLA-LINE      & 10                       & 11.5                      & 12.5                      & 8.8                       & 8.5                       & \textit{89.2}          & 10.26                   \\
                       & RFLA-Triangle  & 16.6                     & 19.7                      & 17.8                      & 14.6                      & 15.9                      & \textit{99.5}          & 16.92                   \\
                       & RFLA-Rectangle & 25.6                     & 29.1                      & 24.4                      & 21.6                      & 25                        & \textit{99.8}          & 25.14                   \\
                       & RFLA-Pengtagon & 25.7                     & 28.8                      & 28.2                      & \textbf{23}               & 24.3                      & \textit{99.8}          & 26                      \\
                       & RFLA-Hextagon  & 26.9                     & 29.3                      & \textbf{30.2}             & 21.8                      & \textbf{25.4}             & \textit{99.8}          & \textbf{26.72}          \\ \hline
\end{tabular}
\end{table*}

Finally, we also evaluate the performance of various attacks under various adversarial defenses.

\end{document}